\documentclass[letterpaper, 10 pt, conference]{ieeeconf}  % Comment this line out if you need a4paper

\IEEEoverridecommandlockouts                              % This command is only needed if 
\usepackage[pdftex]{graphicx}
\usepackage{epstopdf}
\usepackage[cmex10]{amsmath}
\usepackage{algorithm}
\usepackage[noend]{algpseudocode}
\usepackage{soul}
\usepackage{array}
\usepackage{comment}
\usepackage{color}
\usepackage{booktabs, multirow} % for borders and merged ranges
\usepackage{soul}% for underlines
\usepackage[table]{xcolor} % for cell colors
\usepackage{changepage,threeparttable} % for wide tables
\usepackage{fontawesome}
\usepackage{cite}
\usepackage{amsfonts}
\usepackage{lipsum}
\usepackage{verbatim}
\usepackage{graphicx}
\usepackage{notoccite}  
\usepackage{hyperref}
\title{\LARGE \bf
Sim-MEES: Modular End-Effector System Grasping Dataset for Mobile Manipulators in Cluttered Environments
}

\author{Juncheng Li$^{1}$, David J. Cappelleri$^{1}$ % <-this % stops a space
\thanks{$^{1}$ J. Li and D. Cappelleri are with the Multi-Scale Robotics \& Automation Lab, School of Mechanical Engineering, Purdue University, West Lafayette, IN USA.
{\tt\small \{li3670, dcappell\}@purdue.edu}}
}

\begin{document}

\maketitle

%\thispagestyle{empty}
%\pagestyle{empty}

%%%%%%%%%%%%%%%%%%%%%%%%%%%%%%%%%%%%%%%%%%%%%%%%%%%%%%%%%%%%%%%%%%%%%%%%%%%%%%%%
\begin{abstract}
In this paper, we present Sim-MEES: a large-scale synthetic dataset that contains 1,550 objects with varying difficulty levels and physics properties, as well as 11 million grasp labels for mobile manipulators to plan grasps using different gripper modalities in cluttered environments. Our dataset generation process combines analytic models and dynamic simulations of the entire cluttered environment to provide accurate grasp labels. We provide a detailed study of our proposed labeling process for both parallel jaw grippers and suction cup grippers, comparing them with state-of-the-art methods to demonstrate how Sim-MEES can provide precise grasp labels in cluttered environments. 
\end{abstract}

%%%%%%%%%%%%%%%%%%%%%%%%%%%%%%%%%%%%%%%%%%%%%%%%%%%%%%%%%%%%%%%%%%%%%%%%%%%%%%%%
\section{INTRODUCTION}

Robotic grasping is one of the fundamental problems in robotics research. It deals with the task of manipulating objects of different shapes and sizes in various environments along with the selection of appropriate end-effectors. Grasping those objects in cluttered environments~\cite{complex} is a complex task that requires the robotic system to be able to accurately detect and localize the unknown objects, use the end-effectors with appropriate dimensions and mechanisms, as well as to generate robust grasp poses. With the increasing demand for fully autonomous robotic systems to tackle challenging tasks, traditional stationary industrial robotic arms with a single type of end-effector are not suitable for grasping various objects in unstructured and dynamic environments. The development of mobile manipulation platforms with multiple grippers~\cite{TRI,pal_robotics_2023} gains increasing attention, as they can be more effective in performing grasping tasks in such environments. 

\begin{figure}[h!]
\centering
\includegraphics[width=0.8\linewidth]{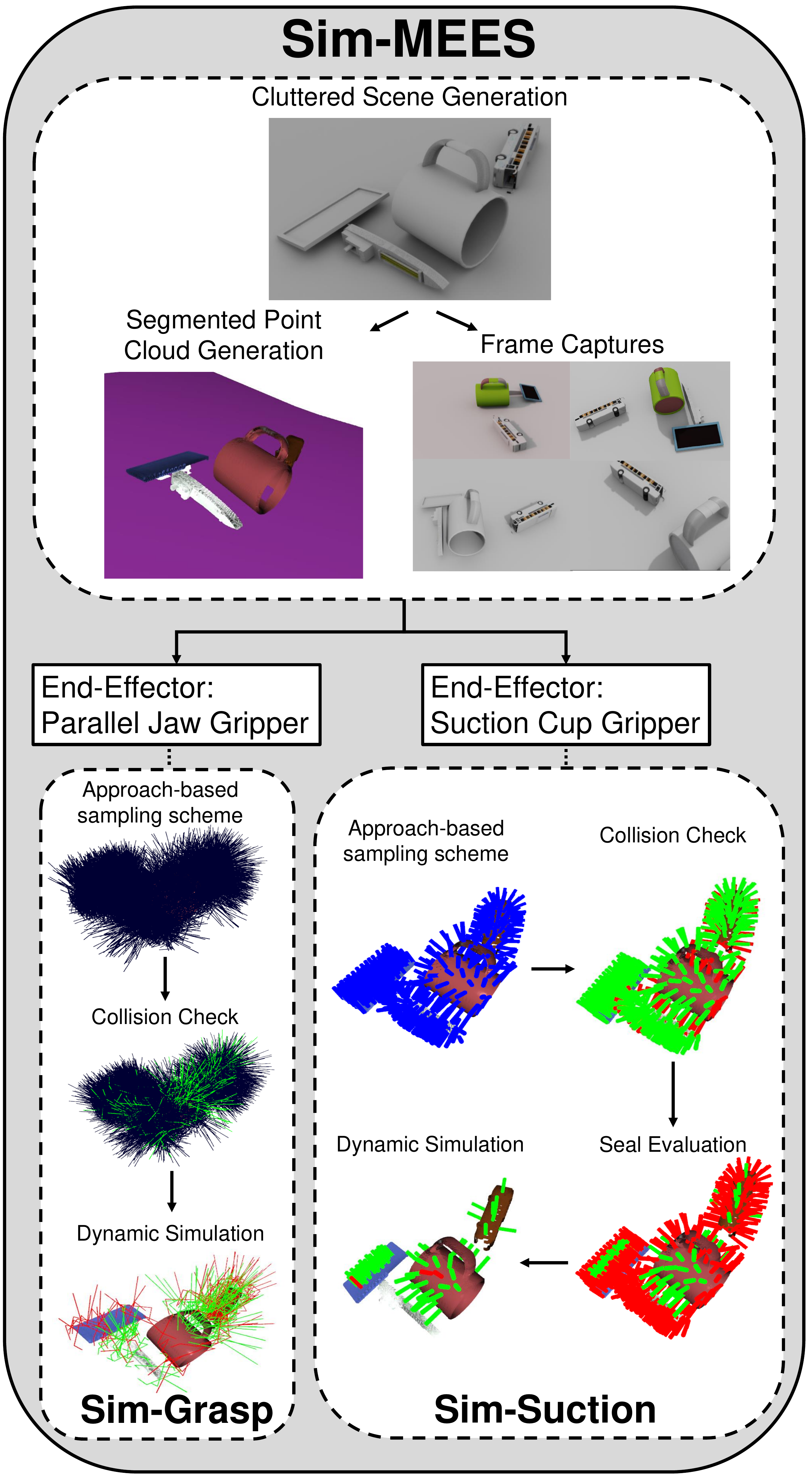}
\vspace{-0.15in}
\caption{\footnotesize The Sim-MEES pipeline involves a synthetic dataset designed for modular end-effector systems operating in cluttered environments, called Sim-MEES~\cite{modular_end_effector}. This dataset includes two modalities of end-effectors, namely parallel jaw grippers (Sim-Grasp) and suction cup grippers (Sim-Suction). The grasps and labels are generated using analytic models and physics simulation.}
\label{pipeline}
\vspace{-0.325in}
\end{figure}
 
Two of the most commonly used end-effectors are the parallel jaw grippers and the suction cup grippers. They are used to pick up and hold objects in tasks such as assembly, material handling, bin picking, and sorting. Parallel jaw grippers are relatively simple to use and can be easily actuated. The high-force closure of parallel jaws can provide a strong grip. The disadvantage is that they are not suitable for handling fragile or soft objects since they can cause some damage to the objects. In addition, they are limited in their ability to handle objects of various sizes in cluttered environments, due to the small jaw opening width, and their overall  size that can collide with the surroundings. Suction cup grippers are another type of end-effector used in robotic applications that are actuated by the vacuum pump to create an airtight seal between a rubber suction cup and the surface, allowing the object to be securely grasped. They are capable of handling objects with irregular surfaces and textures, including those with curved surfaces. However, they are unable to handle objects with uneven surfaces. Both the parallel jaw gripper and the suction cup gripper have their own advantages and disadvantages. Our previous work~\cite{modular_end_effector} proposed a modular end-effector system that enables a mobile manipulation platform to swap between a parallel jaw gripper and a suction cup gripper, making them ideal for use in a variety of applications. The mobility of the platforms allows for great flexibility, so the mobile manipulation platforms can move to different locations to find and grasp different objects~\cite{neural_mobile}. The ability to swap between end-effectors can enable mobile manipulation platforms to handle a wide range of objects and materials. When selecting an end-effector for a particular application, it is important to consider the type of object to be handled, as well as the end-effector's reliability in providing a successful grasp. By understanding when to use each gripper, robots can perform more efficiently and effectively. 

In this paper, we present a novel dataset for a \textbf{M}odular \textbf{E}nd-\textbf{E}ffector \textbf{S}ystem called \textbf{\textit{Sim-MEES}} (see Fig.~\ref{pipeline}) that utilizes the parallel jaw gripper and the suction cup gripper. The dimension and physics constraints of selected grippers can be easily adjusted. We select the Fetch robot gripper~\cite{FetchRobotics2} and the Robotiq 2F-140 gripper~\cite{robotiq} as our parallel jaw grippers, along with a 1.5 cm radius bellows suction cup and a 2.5 cm radius bellows suction cup for our suction cup gripper. The dataset contains a total of $11$ million parallel jaw grasps and suction grasps generated from $1,550$ objects in $500$ cluttered synthetic environments, captured from $400,000$ randomly sampled camera viewpoints. The objects come from ShapeNet~\cite{shapenet}, YCB objects~\cite{ycb}, NVIDIA Omniverse Assets~\cite{nvidia}, and Adversarial Objects in Dex-Net~\cite{dexnet1}. We propose a pipeline to study the grasp correlation of each object in the entire cluttered environment in order to automate the generation of accurately labeled data. This is achieved by combining sampling-based approaches, analytical model analysis, domain randomization, and dynamic physics simulations. Each grasp has a corresponding 6-D grasp pose, gripper dimension, object difficulty level, object physics information, RGB-D image, camera matrix, binary success label, and object segmentation point cloud associated with it. This comprehensive dataset and auto-labeling pipeline are intended to serve as a synthetic benchmark and reference for grasping research. It provides a common dataset and approach for comparison and evaluation of grasping algorithms across different modality grippers, and can also be used to develop sim-to-real data-driven methods of multiple grippers grasping in real-world scenarios.

%%%%%%%%%%%%%%%%%%%%%%%%%%%%%%%%%%%%%%%%%%%%%%%%%%%%%%%%%%%%%%%%%%%%%%%%%%%%%%%%
\section{RELATED WORK}
Data-driven approaches~\cite{datadriven1,datadriven2,datadriven3,datadriven4,data-driven-review} gained popularity in recent years as a method for tackling grasp detection problems. While traditional approaches rely on object geometries ~\cite{gpd,traditional1,traditional2,traditional3,traditional4,traditional5,traditional6,traditional7,traditional8} and have achieved some success, they are limited by slow operating speeds, high-resolution sensor requirements, and poor generalization. In contrast, data-driven approaches learn from vast amounts of data containing information such as object shape, size, weight, and surface texture, allowing them to achieve real-time performance and improve grasp detection accuracy. However, obtaining such a large-scale dataset with accurately annotated ground truth data can be difficult. The empirical human labeling process~\cite{zeng} requires experienced humans to manually annotate each pixel in the RGB-D images as graspable areas, which is time-consuming and cannot be easily used to generate very large datasets. The process of generating real-world datasets~\cite{graspnet},\cite{real_robot} experiences the same difficulties, and it is also challenging to generalize to different environments and vision (camera) settings. Therefore, the real-world dataset generation process is not well suitable for mobile manipulation platforms. While much research has been conducted to improve the performance of one gripper type~\cite{jiang,billion,PointNetGPD,dexnet3,object_agnostic_suction,end-to-end}, such as parallel jaw grippers or suction cup grippers, there is a lack of studies~\cite{dexnet4.0,zeng,graspnet,suctionnet} that investigate the
effects of combining different gripper types or different gripper dimensions in a single grasping task. A comprehensive dataset and pipeline containing multiple gripper types and dimensions would not only allow researchers to evaluate the effectiveness of different combinations of gripper types but could also provide insight into the best practices for the design and
utilization of multiple grippers for robotic grasping. 

We compare \textit{Sim-MEES} with other publicly available grasp datasets and summarize the main differences in Table~\ref{table:Comparison of the suction grasp dataset}. 
\begin{table*}[h]
\caption{Comparison of Grasping Datasets.}\label{table:Comparison of the suction grasp dataset}
\vspace{-0.10in}
\scalebox{0.95}{
\begin{tabular}{@{}cccccccccc@{}}
\toprule
\multirow{2}{*}{\textbf{Dataset}} & \textbf{Grasp Pose}     & \textbf{Objects/} & \textbf{Camera} & \textbf{Total}   & \textbf{Total}  & \textbf{Gripper}       & \textbf{Multiple Gripper} & \textbf{Semantic}     & \textbf{Dynamics}   \\
                                  & \textbf{Label (Method)} & \textbf{Scene}    & \textbf{Type}   & \textbf{Objects} & \textbf{Labels} & \textbf{Modality}      & \textbf{sizes}            & \textbf{Segmentation} & \textbf{Evaluation} \\ \midrule
Cornell~\cite{cornell}                           & 2D (\faUser)                  & 1                 & Real            & 240              & 8K              & PGrip.                 & Yes                       & No                    & No                  \\
Jacquard~\cite{Jacquard}                          & 2D (\faPlay )                  & 1                 & Sim             & 11K              & 1.1M            & PGrip.                 & Yes                       & No                    & Partial             \\
ACRONYM~\cite{ACRONYM}                           & 2D (\faPlay )                  & 1                 & Sim             & 8.8K             & 17.7M           & PGrip.                 & No                        & Yes                   & Partial             \\
6-DOF GraspNet~\cite{6dofgraspnet}                   & 6D (\faPlay )                  & 1                 & Sim             & 206              & 7.07M           & PGrip.                 & No                        & Yes                   & Partial             \\
GraspNet~\cite{graspnet}                          & 6D (\faPencil )                  & $\sim$10          & Real            & 88               & $\sim$1.2B      & PGrip.                 & No                        & Yes                   & No                  \\
SuctionNet~\cite{suctionnet}                       & 6D (\faPencil )                  & $\sim$10          & Real            & 88               & $\sim$1.1B      & SGrip.                 & No                        & Yes                   & No                  \\
Dex-Net 4.0~\cite{dexnet4.0}                       & 2D (\faPencil )                  & 3-10              & Sim             & 1.6K             & $5$M            & PGrip.+SGrip.          & No                        & No                    & No                  \\
A. Zeng~\cite{zeng}                            & 2D (\faUser )                  & NA                & Real            & NA               & 191M            & PGrip.+SGrip.          & No                        & No                    & No                  \\
\textit{\textbf{Sim-MEES}}        & \textbf{6D (\faPencil , \faPlay )}      & \textbf{1-20}     & \textbf{Sim}    & \textbf{1.5K}    & \textbf{~11M}   & \textbf{PGrip.+SGrip.} & \textbf{Yes}              & \textbf{Yes}          & \textbf{Yes}        \\ \bottomrule
\end{tabular}
}
\begin{tablenotes}
      %\small
      \item Note: PGrip = Parallel jaw gripper, SGrip = suction cup gripper.  Grasp labels can be generated either manually (\faUser), using analytical models (\faPencil), or through physics simulation (\faPlay). Dynamics evaluation is denoted as partial when it only evaluates an isolated single object, rather than objects in cluttered environments.
    \end{tablenotes}
    \vspace{-0.20in}
\end{table*}
The comparison reveals that our dataset employs analytical models, physics simulation, and comprehensive dynamics evaluation to enable an automated generation process that accurately annotates large-scale 6D grasp pose labels for various sizes of parallel jaw grippers and suction cup grippers. In~\cite{ACRONYM}, it is noted that performing partial dynamic evaluation on a single isolated object and using that information to label cluttered environments could result in false negatives and false positives due to the significant impact of environmental contacts and interactions.
The state-of-art dataset most similar to our work is Dex-Net 4.0~\cite{dexnet4.0}. It contains 5 million grasps with parallel jaw gripper and suction cup gripper associated with synthetic point cloud and grasp metrics computed from 1664 unique 3D objects in simulated heaps. This large synthetic dataset enables the training of a neural network to detect grasps in images and prove the effectiveness of learning from synthetic dataset to perform well in real-world grasping. However, this dataset and its potential applications has some limitations. First, this dataset depth image is only captured from a fixed overhead single-viewed simulated depth camera. It does not have any variation in the viewpoint and occlusions. The labeled ground truth grasp pose for the parallel jaw grippers and suction cup grippers are 3D. According to the author's talk in the seminar~\cite{talk}, Dex-Net 4.0 is unable to solve the challenges in mobile manipulation platforms with a moving camera that is not mounted on top of the workspace. Secondly, the analytical model proposed by the author only takes into account the resistance of the wrench when grasping a single object, and fails to consider the cluttered environment as a whole. This is a very common failure scenario in cluttered environments when a robot attempts to grasp an object from the bottom of a pile with other objects stacked on top of it. Furthermore, the analytical model overlooks cases where objects are in close proximity to each other because the dataset lacks object segmentation information. Thirdly, the analytical model for the suction cup gripper neglects to examine scenarios where uneven surfaces and holes underneath the suction cup may lead to air leaks, making it impossible to form a proper seal. We provide a comparative study of Dex-Net 4.0 and our model on challenging corner cases in Section~\ref{section:exp} and explain how our model can tackle the mentioned issues.

%%%%%%%%%%%%%%%%%%%%%%%%%%%%%%%%%%%%%%%%%%%%%%%%%%%%%%%%%%%%%%%%%%%%%%%%%%%%%%%%
\section{Sim-MEES}
We provide methods for generating a large-scale synthetic dataset for modular end-effector systems in cluttered environments that combine analytic models and dynamic interactions. Sim-MEES includes 1550 objects from 137 categories, encompassing a range of grasp difficulty levels, from simple primitive shapes to complex geometries. Our dataset has a wider coverage on object difficult levels compared to ACRONYM~\cite{ACRONYM}, which excludes meshes that contain more than one connected component.  We use Isaac Sim simulator~\cite{isaac} to create cluttered environments and perform physics simulations. We utilize domain randomization by randomly selecting camera viewpoints, lighting conditions, object scales, and material properties to produce 400,000 photo-realistic RGB-D images, each with an accompanying single-view point cloud. Subsequently, we conduct both analytic and dynamic analyses, using various sizes of parallel jaw and suction cup grippers in 500 cluttered environments, to create labeled datasets known as \textbf{\textit{Sim-Grasp}} and \textbf{\textit{Sim-Suction}}.

\subsection{Cluttered Scene Generation}
To create an unstructured environment, we randomly select and drop objects $\mathcal{O}$ onto the ground plane in Isaac Sim Simulator~\cite{isaac}. We define our state distribution $\zeta_{scene}$ as a product of the following factors:
\begin{itemize}
  \item Total objects number $\mathrm{I}$ in one scene: Randomly selected from a range $[1,20]$.
  \item Objects $\mathcal{O}$: uniformly randomly sampled with replacement of total size $\mathrm{I}$ from 1550 objects pool. Objects $\mathcal{O}$ drop locations are uniformly sampled from the SE(3) space $[-0.1~m,-0.1~m,0.5~m] \times [0.1~m,0.1~m,0.8~m]$ above the ground plane. Objects $\mathcal{O}$'s orientations are uniformly randomly sampled. Each $\mathcal{O}_i$ is dropped at free fall with mass $\mathrm{m}_i$.
  \item Coulomb friction coefficient $\mu$: Randomly sampled from the range $[0,1]$ and used to model tangential forces between contact surfaces.
 \item Object scale: randomly sampled from the range $[1,1.5]$.
 \item Object mass: auto-generated by using material density multiplied by mesh volume.
 
 \end{itemize} 
 
  For the simulator to implement multi-body physics, we use high-detail triangle meshes to describe the collision geometry, where several convex shapes approximate the input mesh. We first sample the initial state from the state distribution $\zeta_{scene}$ to begin the data generation process. Then we start the dynamics simulation and drop objects one-by-one with a rendering time step of $0.5~s$ to avoid object penetration until all objects on the ground plane reach their static equilibrium. Finally, we repeat this process to populate the dataset with $500$ cluttered scenes and save the object poses and their semantic instance names.

\subsection{Frame Observation Generation}

%\begin{figure}
%\centering
%\includegraphics[width=0.8\linewidth]{sim-mees/camera_location.pdf}
%\vspace{-0.10in}
%\caption{\footnotesize Illustration of viewpoint locations in sphere coordinate system. Radius $\rho$ varies from $0.5~m$ to $10~m$. The blue arrows represent the sampled camera configurations at $\rho=0.5~m$. Different lighting conditions aim to simulate real-world lighting and address the domain gap problem. }
%\label{camera_location}

%\end{figure}

\begin{figure}
\centering
\includegraphics[width=1\linewidth]{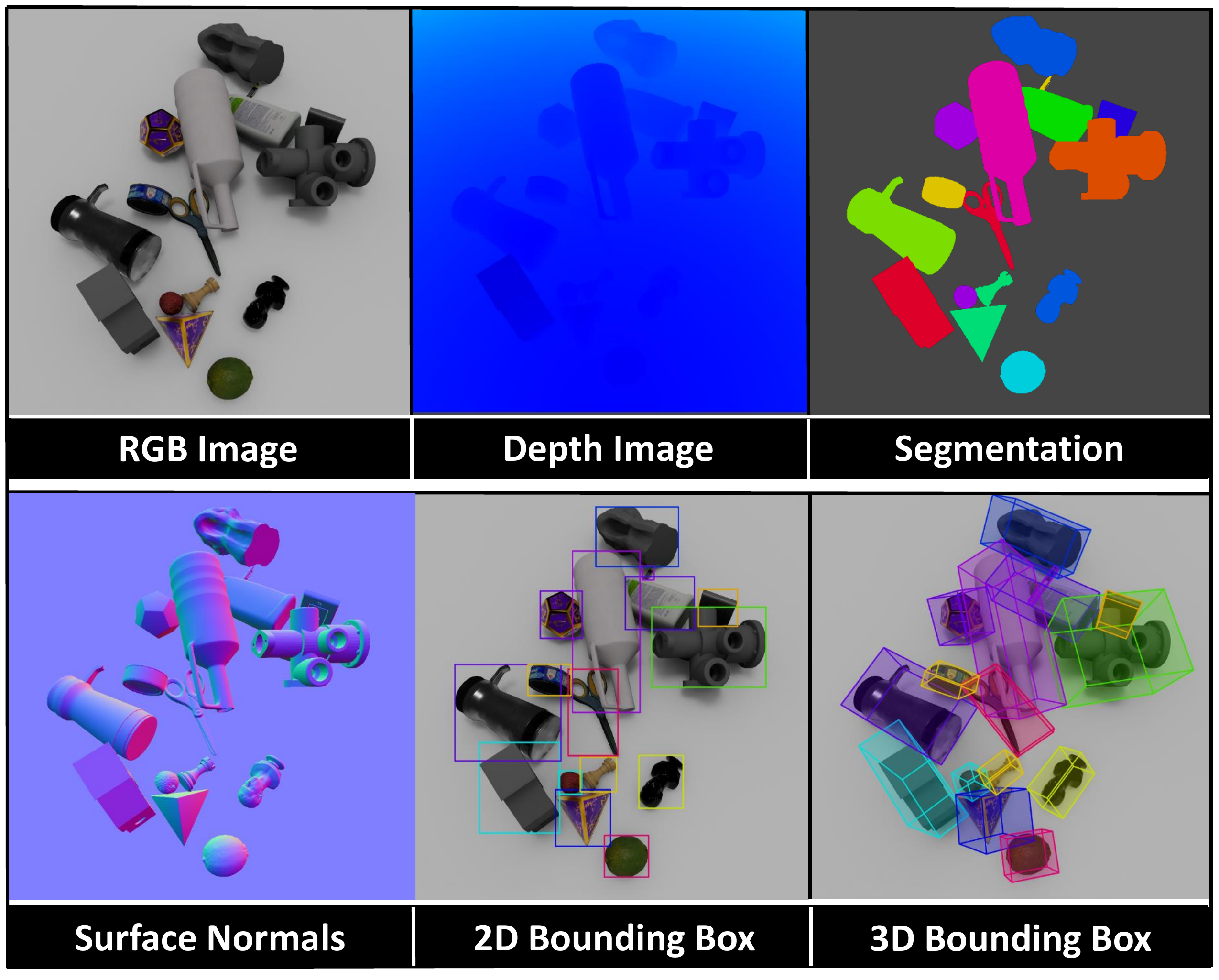}
\vspace{-0.25in}
\caption{\footnotesize Key components of our \textit{Sim-MEES}. The Photo-realistic RGB-D images are rendered from a synthetic camera with an intrinsic matrix sampled around the nominal values of a PrimeSense Carmine $1.09$ camera. The segmentation mask is generated using the GPU-RayTracing in PhysX engine~\cite{tracing}. The segmented point cloud can be registered from 2D RGB-D images and surface normals with the help of a segmentation mask using camera intrinsic and extrinsic matrices.
%The Annotated 6D suction poses are shown in Fig.~\ref{annotate}. 
We also provide the 2D and 3D bounding box labels for each object instance, which can contribute to the object detection and pose estimation community.}
\label{frame}
\vspace{-0.20in}
\end{figure}

With $500$ cluttered scenes, we further populate the dataset by capturing $800$ frames from viewpoint $\mathcal{V}$ for each scene, resulting in $400,000$ frames in total. We define the viewpoint distribution $\zeta_{viewpoint}$ as a product of the following:
\begin{itemize}

\item Viewpoint pose: positions are uniformly sampled from the sphere coordinate space $[\rho=0.5~m,\phi=0,\theta=0] \times [\rho=10~m,\phi=\pi/2,\theta=\pi]$ with the orientation pointing towards the scene center (0,0,0).
\item Lighting condition: the location of the sphere-shaped light source is uniformly sampled from the 3D space $[-4.5~m,-4.5~m,3~m] \times [4.5~m,4.5~m,8~m]$.
The radius of the sphere light source is uniformly randomly sampled from $[0.5~m,2~m]$. The light intensity is uniformly randomly sampled from $[3000~Lux,20000~Lux]$. The color of the light is uniformly randomly sampled from a list of common light color, from warehouse environments.  
\end{itemize}
We draw each frame's viewpoint state from the viewpoint distribution $\zeta_{viewpoint}$ and record the synchronized RGB image, depth information, 2D bounding boxes, 3D bounding boxes, instance segmentation mask, semantic segmentation mask, and camera matrix as shown in Fig.\ref{frame}. 

\subsection{Sim-Grasp}
Sim-Grasp uses two parallel jaw grippers (Fig.~\ref{model}), namely, the Fetch gripper and Robotiq gripper. 
\subsubsection{Grasp Sampling}

To generate grasp candidates, we use an approach-based sampling scheme. We first employ iterative Farthest Point Sampling (FPS)~\cite{FPS} to select a set of points $\mathbb{T}$ from each object point cloud. This method is known for its superior coverage of the object surface when compared to random sampling. For each sampled grasp point $t$ on the differentiable object surface, we calculate the corresponding Darboux frame, denoted as $\mathrm{R}\in \mathbb{R}$. A Darboux frame is a natural moving frame constructed on a surface for the purpose of studying curves:

\begin{equation}
R(t)=[v_1(t)|v_2(t)|v_3(t)],
\label{equ:R(t)}
\end{equation}
where $v_1(t)\in \mathcal{N}$ is the normal vector, $v_2(t)$ is the major axis of curvature vector, and $v_3(t)$ is the minor axis of curvature vector. To calculate these vectors, we evaluate the Eigenvectors of the $3\times3$ matrix $N(t)$ given by:
\begin{equation}
N(t)=\sum_{t\in\mathbb{T}}^{}\hat{n}(t)\hat{n}^T(t),
\end{equation}

where $\hat{n}(t)$ is the normal vector at point t. 
The Eigenvectors of matrix $N(t)$ in decreasing order are denoted as $[v_3(t),v_2(t),v_1(t)]$. We align the $X$-axis of our grasp candidates with $v_1(t)$, the $Y$-axis with $v_2(t)$, and the $Z$-axis with $v_3(t)$. If $v_1(t)$ points in the opposite direction to $\hat{n}(t)$, we rotate the matrix $R(t)$ along the z-axis by $180$ degrees to ensure that it points outward from the object surface. We then uniformly rotate the parallel jaw gripper around the approach vector $v_1(t)$ between 0 and 2$\pi$. We choose the sampling parameters $(\alpha,\beta)$ to be $(\pi/2,0)$, as this has been found to provide better robust coverage according to a study in~\cite{billion}. We repeat this process for every point $t\in\mathbb{T}$ to obtain a set of grasp candidates for each cluttered scene, along with the corresponding object instance and 6D pose information.

\subsubsection{Grasp Evaluation}
To check for possible collisions of the gripper with the environment in Isaac Sim simulator~\cite{isaac}, we utilize the Overlap query by providing the triangle meshes of the gripper models. Additionally, we apply a commonly used heuristic that checks whether the close-up region of the gripper's finger is empty. After obtaining grasp candidates that pass the collision check, we evaluate and label each grasp using Isaac Sim. To provide realistic physics simulation, we mount the gripper on a 7-DoF UR10 robot and use the RMPflow controller~\cite{RMPFLOW} to simulate the dynamic behavior of the gripper. 
The grippers are controlled with articulation controllers. To close the fingers, we use a force threshold. Each object in the cluttered environment is modeled as a rigid object with mass and physics material properties, enabling us to simulate the interactions between each object. However, previous works~\cite{ACRONYM,graspnet,dexnet4.0} have used the results achieved from single object simulation to label the cluttered environment. This simplification method has led to many inaccurate ground-truth labels.

\subsection{Sim-Suction}
Sim-Suction uses 1.5cm and 2.5cm radius bellows suction cup grippers.
\subsubsection{Suction Sampling}
We use a similar sampling scheme as in the case of Sim-Grasp. To form a seal, we want to align the suction gripper approaching vector with the objects’ surface normal on a sampled suction grasp point, therefore we only apply constraints on the gripper's X-axis.
\subsubsection{Suction Evaluation}
\begin{figure}%[h]
\centering
\includegraphics[width=1\linewidth]{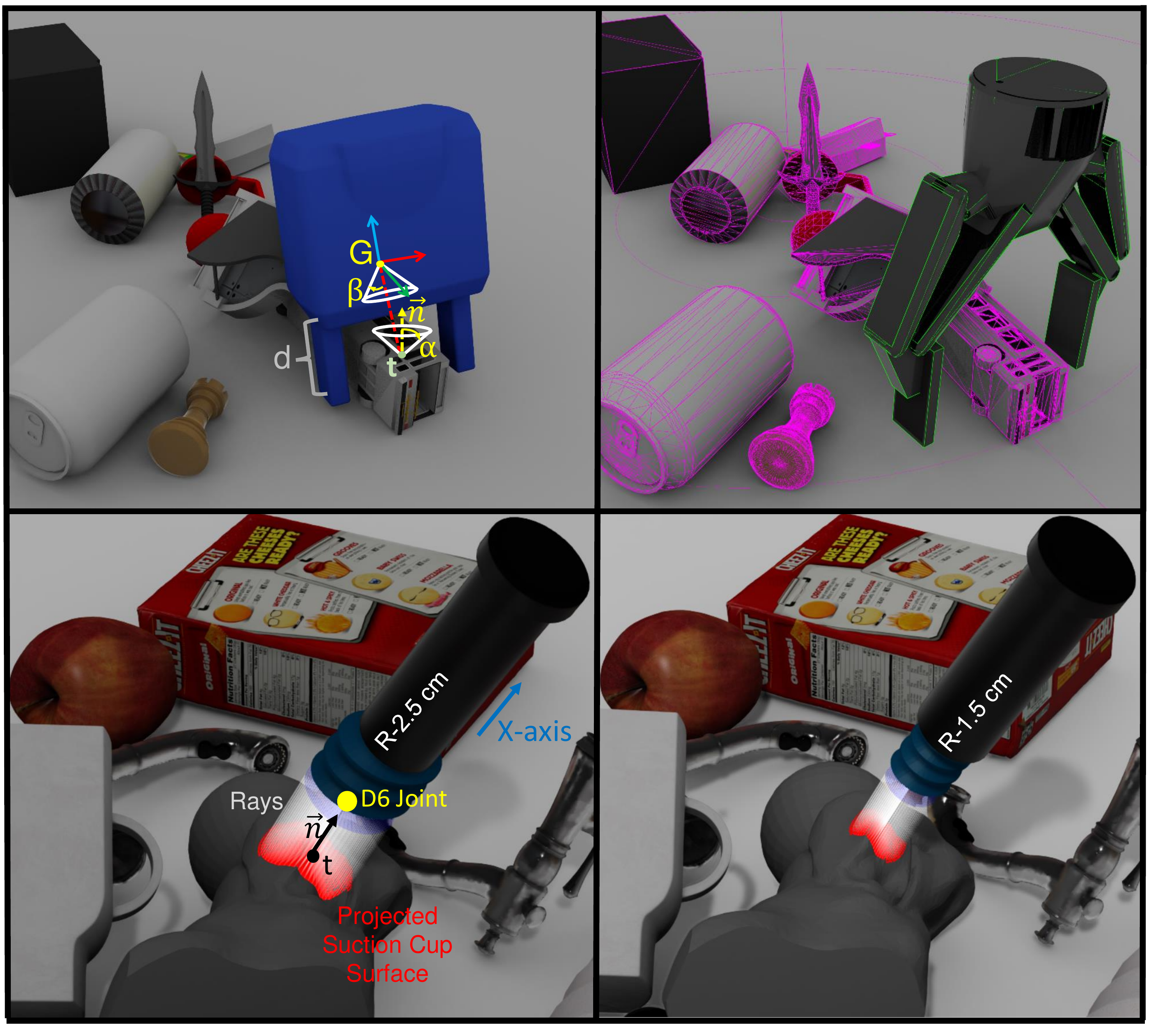}
\vspace{-0.20in}
\caption{\footnotesize \textbf{Top-Left (Fetch gripper).} Parameterization of the approach-based grasp sampling schemes (RGB: YXZ). We uniformly sample the gripper standoff from 0 to $d$. \textbf{Top-Right (Robotiq gripper).} We use triangle mesh to accurately define object collision. \textbf{Bottom-Left (2.5 cm radius bellows suction cup).} We evaluate the seal performance by casting dense rays along surface normal vectors from the suction cup surface towards the object surface. To evaluate the suction dynamics, we model the suction cup gripper with a 6 degree of freedom joint. We set the suction cup bending angle limit to lock individual axes. We set $30$ N force limit for $2.5$ cm suction cup and check if the 6D joint can be created and maintained during the manipulator movement. \textbf{Bottom-Right (1.5 cm radius bellows suction cup).} We set the $20$ N force limit for $1.5$ cm suction cup.}
\label{model}
\vspace{-0.10in}
\end{figure}

Suction cup grippers can lift an object when the pressure difference between the atmosphere and the vacuum is large enough. Intuitively, a suction cup gripper can easily form an airtight seal on a flat surface. However, forming a seal with the suction cup becomes a challenge when dealing with an irregular surface. To address this issue, we took inspiration from a spider's web and model the bellows suction cup with 15 concentric polygons, each with 64 vertices. We perform the collision check in a physics simulator by casting rays along the x-axis from each vertex, as shown in the Fig.~\ref{model}, to detect the closest object that intersects with a specified ray. We evaluate the suction cup's seal by modeling it with deformable material and as a spring, with a deformable threshold of 10\%. Our analytical suction cup model can be utilized for assessing suction seals on intricate geometries, including uneven surfaces and surfaces with holes or grooves. In addition to analytical analysis, we also conduct a dynamic evaluation of suction. We use a 6D (6-DoF) joint to represent the suction cup mechanism and evaluate suction dynamics to determine whether the suction cup can establish and sustain the 6D joint between the object surface and the suction cup surface during movement. For a 1.5 cm suction cup, we set the force limit at 20 N, and for a 2.5 cm suction cup, we set the force limit at 30 N. We use a bending angle threshold of 7 degrees for both grippers. The gripper is also attached to a manipulator to better get realistic physics. 

%\section{Sim-Suction}
%\input{sim-suction.tex}
%%%%%%%%%%%%%%%%%%%%%%%%%%%%%%%%%%%%%%%%%%%%%%%%%%%%%%%%%%%%%%%%%%%%%%%%%%%%%%%%
\section{EXPERIMENTS}
\label{section:exp}

\subsection{Object Diversity}
\begin{figure}
\centering
\includegraphics[width=1\linewidth]{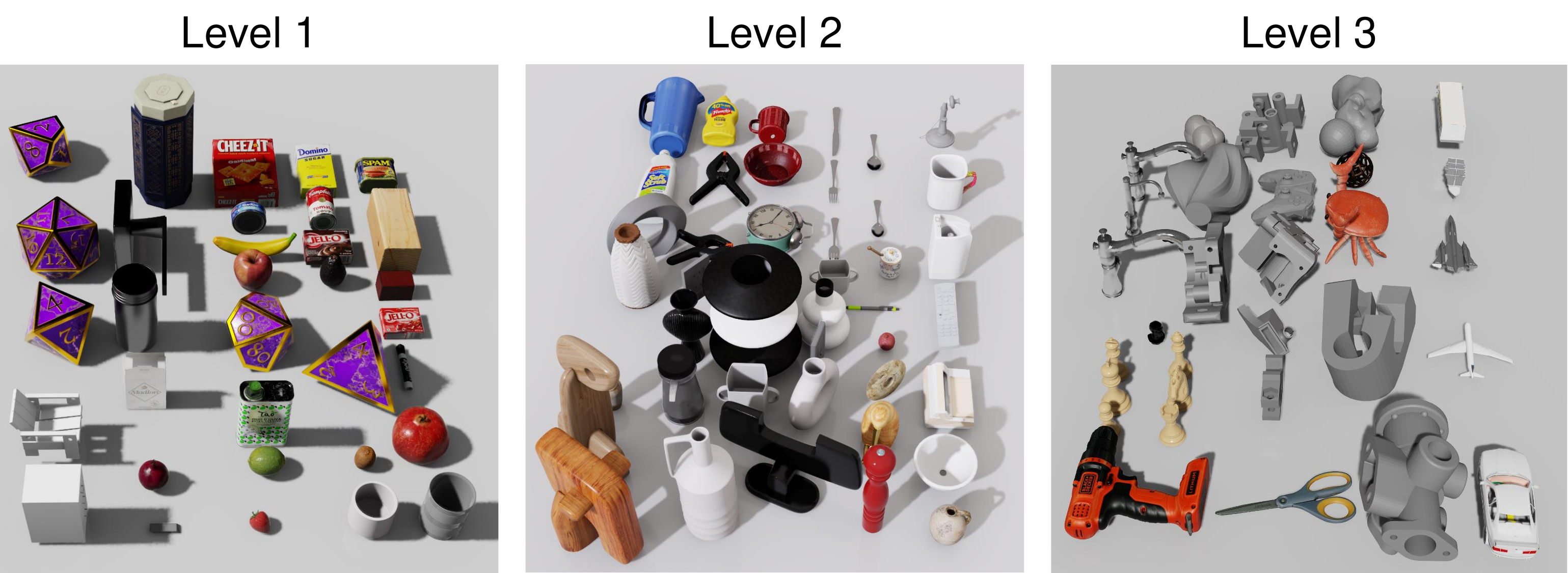}
\vspace{-0.20in}
\caption{\footnotesize Example of objects with different difficulty levels. Level 1 is the least difficult; Level 3 is the most difficult.}
\label{level}
\vspace{-0.20in}
\end{figure}

\begin{figure}
\centering
\includegraphics[width=1.0\linewidth]{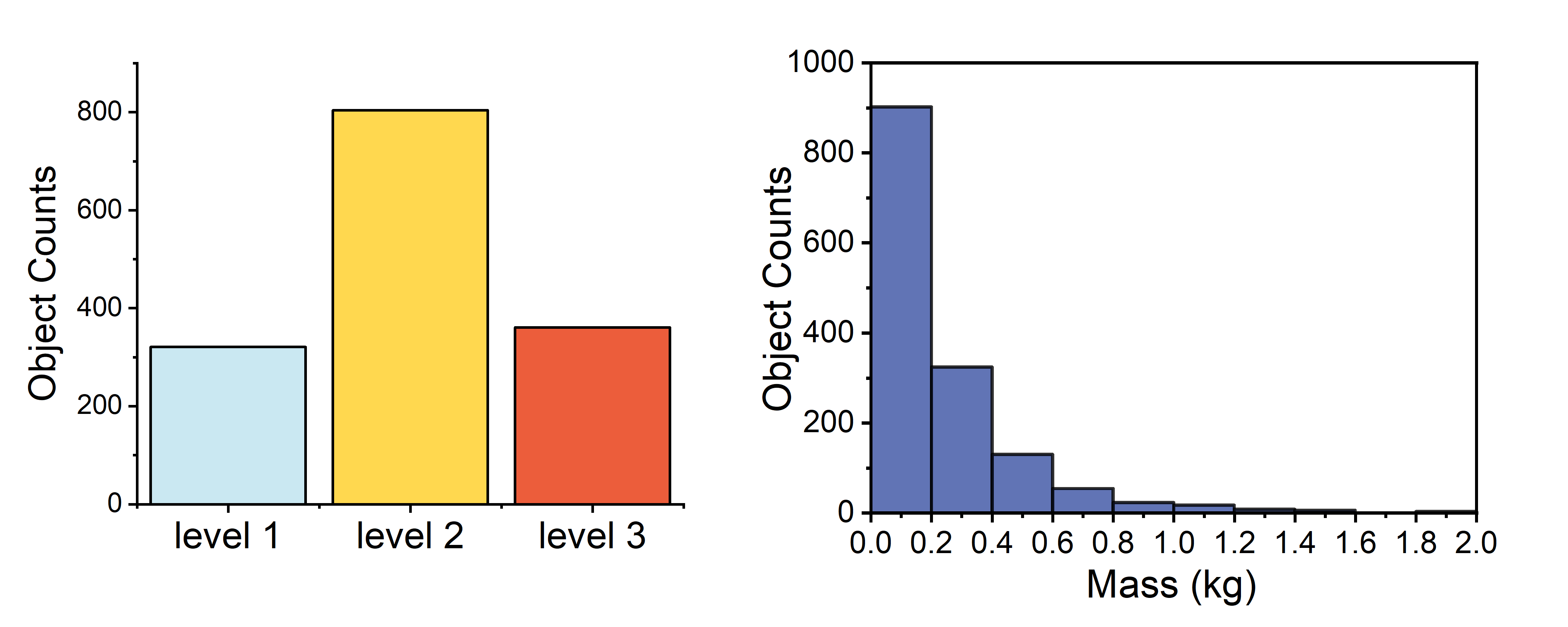}
\vspace{-0.30in}
\caption{\footnotesize Distribution of object mass and difficulty levels for objects in SIM-MEES dataset.}
\label{mass}
\vspace{-0.30in}
\end{figure}

Sim-MEES contains a wide range of object types categorized into three difficulty levels (Fig.~\ref{level}): Level 1 includes prismatic and circular solids, Level 2 includes objects with varied geometry, and Level 3 includes objects with adversarial geometry and material properties. The resulting mass and difficulty level distribution are shown in Fig.~\ref{mass}. The difficulty levels of the objects are determined based on the complexity of their triangle mesh. We use the material density of each object to calculate the object's mass. 

\begin{figure*}
\centering
\includegraphics[width=0.88\linewidth]{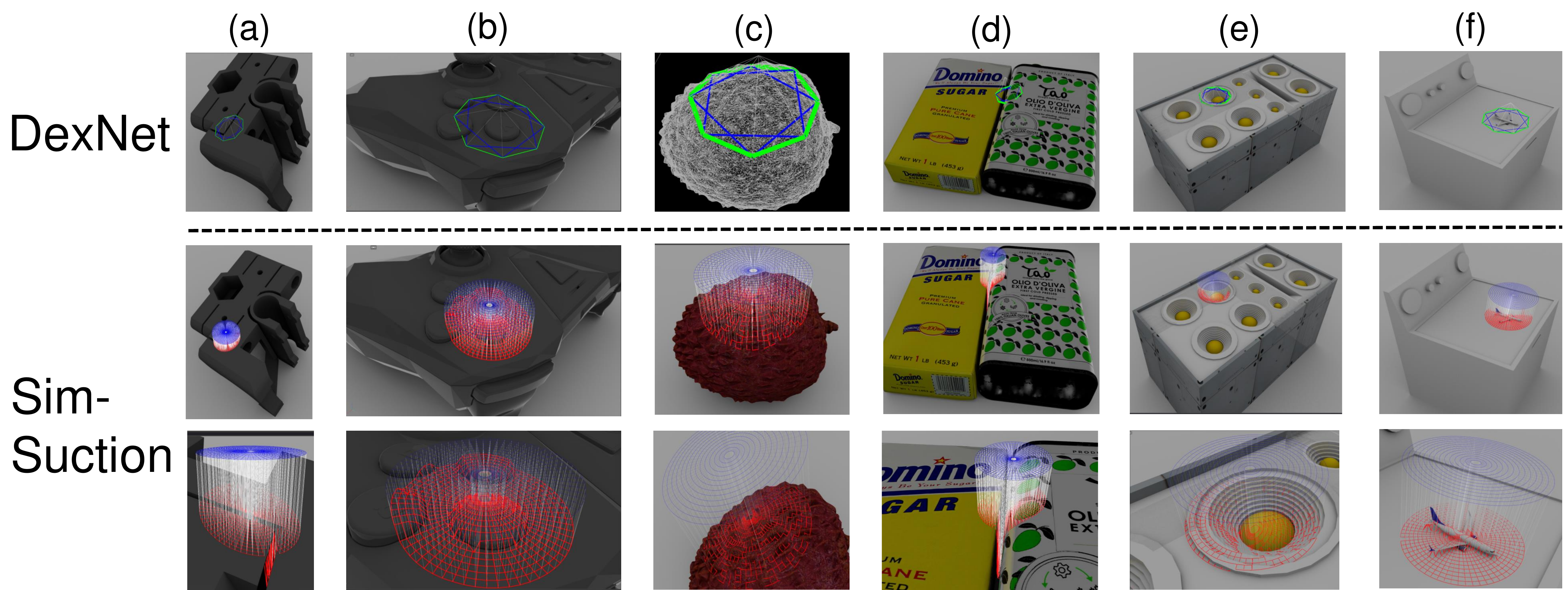}
\vspace{-0.10in}
\caption{\footnotesize Corner cases for 1.5 cm suction cup gripper seal evaluation. (a) Surface with grooves and holes. (b) Protruding parts. (c) Rough surface. (d) Objects next to each other. (e) Concave surfaces. (f) Overlapped objects.}
\label{compare-suction}
\vspace{-0.20in}
\end{figure*}

\subsection{Annotation Evaluation}

Robots equipped with suction cup grippers and parallel jaw grippers that physically interact with objects in cluttered environments face inherent uncertainties regarding how objects will react to grasping. Previous methods~\cite{suctionnet,graspnet,dexnet4.0} have failed to accurately consider the collision geometry of the gripper, relying only on simplified cubes and cylinders for collision checks. Furthermore, these methods have not taken into account the correlation between the object of interest and its surroundings, evaluating grasp candidates only on the isolated object. We believe that such methods decrease the accuracy of ground truth labeling by introducing false positives. 

We evaluated previous labeling approaches on our dataset and compared each evaluation pass result with our methods to provide an ablation study, as shown in the Table~\ref{table:annotation}. The pipeline has three ordered sub-evaluation sequences with binary-valued metric (0 is fail, 1 is pass) on each suction candidate $\mathrm{S}$: Collision check $\mathcal{Q}_{collision}(\mathrm{S})=\{0,1\}$, Seal formation evaluation $\mathcal{Q}_{seal}(\mathrm{S})=\{0,1\}$, and dynamics evaluation $\mathcal{Q}_{dynamics}(\mathrm{S})=\{0,1\}$. We calculate the collision check passing rate by using the number of candidates with $\mathcal{Q}_{collision}=1$ divided by the total number of sampled candidates. We calculate the seal check passing rate by using the number of candidates with $\mathcal{Q}_{collision}(\mathrm{S}) \times\mathcal{Q}_{seal}(\mathrm{S})=1$ divided by the total number of candidates with $\mathcal{Q}_{collision}(\mathrm{S})=1$. We calculate the dynamic check passing rate by using the number of candidates with $ \mathcal{Q}_{collision}(\mathrm{S})\times\mathcal{Q}_{seal}(\mathrm{S})\times\mathcal{Q}_{dynamics}(\mathrm{S})=1$ divided by the total number of candidates with $\mathcal{Q}_{collision}(\mathrm{S}) \times\mathcal{Q}_{seal}(\mathrm{S})=1$. We investigated the impact of individual approaches on the pass rate of three evaluation checks. To ensure fairness, we only altered the approach we were interested in and kept all other steps the same as ours, as illustrated in Table~\ref{table:annotation}. Specifically, Sim-Suction-DexNetModel uses an 8-vertex quasi-static spring suction model presented in~\cite{dexnet3}, while Sim-Suction utilizes a 960-vertex model. Our results show that Sim-Suction-DexNetModel has a higher pass rate for the seal evaluation check compared to Sim-Suction. This is due to the fact that the suction model from DexNet is not capable of evaluating seal performance on complex geometries, which results in more false positives. Furthermore, only a portion of these false positives are rejected during dynamic evaluation. Sim-Suction-SingleObjectDynamics and Sim-Grasp-SingleObjectDynamics focus solely on the dynamic behavior of the single object being grasped, disregarding any interactions with the environment. As a result, these approaches achieve higher passing rates during dynamic evaluation compared to Sim-Suction and Sim-Grasp. However, these passing rate differences can be attributed to the ability of Sim-Suction and Sim-Grasp to reject false positives by considering the dynamics of cluttered environments. Sim-Grasp-Simplified-Gripper employs primitive cubes and cylinders to represent the geometry of the gripper mesh. While this method is unlikely to have an impact on the sampling scheme when the gripper depth and open width are correctly defined, it can result in false positives during the collision check process, particularly in highly cluttered environments.
\begin{table}[]
\caption{Ablation study of the labeling methods.}
\label{table:annotation}
\vspace{-0.10in}
\scalebox{0.7}{
\begin{tabular}{|c|c|c|c|c|}
\hline
\multirow{2}{*}{\textbf{Methods}}                                          & \textbf{Collision Check}          & \textbf{Seal Evaluation}          & \textbf{Simulation Evaluation}    & \textbf{False}                \\
                                                                           & \textbf{Passing Rate}             & \textbf{Passing  Rate}            & \textbf{Passing Rate}             & \textbf{Positives}            \\ \hline
\begin{tabular}[c]{@{}c@{}}Sim-Suction-\\ DexNetModel\end{tabular}         & 57.17\%                           & 28.26\%                           & 84.37\%                            & High                          \\ \hline
\begin{tabular}[c]{@{}c@{}}Sim-Suction-\\ SingleObjectDynamic\end{tabular} & 57.83\%                           & 14.73\%                            & 95.42\%                            & High                          \\ \hline
\multirow{2}{*}{\textbf{Sim-Suction}}                                      & \multirow{2}{*}{\textbf{58.05\%}} & \multirow{2}{*}{\textbf{14.11\%}} & \multirow{2}{*}{\textbf{88.73\%}} & \multirow{2}{*}{\textbf{Low}} \\
                                                                           &                                   &                                   &                                   &                               \\ \hline
\begin{tabular}[c]{@{}c@{}}Sim-Grasp-\\ Simplified-Gripper\end{tabular}    & 21.11\%                           & -                                 & 12.34\%                           & High                          \\ \hline
\begin{tabular}[c]{@{}c@{}}Sim-Grasp-\\ SingleObjectDynamic\end{tabular}   & 3.42\%                             & -                                 & 77.66\%                            & High                          \\ \hline
\multirow{2}{*}{\textit{\textbf{Sim-Grasp}}}                               & \multirow{2}{*}{\textbf{3.46\%}}  & \multirow{2}{*}{\textbf{-}}       & \multirow{2}{*}{\textbf{51.05\%}}    & \multirow{2}{*}{\textbf{Low}} \\
                                                                           &                                   &                                   &                                   &                               \\ \hline
\end{tabular}
}
\begin{tablenotes}
      %\small
      \item Note: Fetch parallel jaw gripper and 1.5 cm suction cup gripper used for experiments. 
    \end{tablenotes}
    \vspace{-0.20in}
\end{table}

\subsection{Corner Cases Study}
The results from Table~\ref{table:annotation} show that, with our suction analytical model seal evaluation, Sim-Suction can eliminate 14.15\% of the candidates after the collision check, which are false positives, compared to DexNet. We provide a comprehensive study of those false positive corner cases, as shown in Fig~\ref{compare-suction}. In unstructured environments with different difficulty levels of objects, the ground truth labeling process is expected to handle different scenarios and provide an accurate result. However, the previous suction model proposed by Dex-Net and used by others ~\cite{suctionnet, zhang} has limitations in dealing with complex geometries and overlapped environments. The DexNet model utilizes the perimeter, flexion, and cone spring connected by eight vertices to assess the seal formation. These vertices are selected on the outer perimeter, resulting in the neglect of any geometry inside the suction cup perimeter. Theoretically, if the suction cup gripper's radius is small enough, and the object is non-porous, there would be no need to be concerned about any geometry inside the suction cup causing air leaks. However, in reality, suction cup grippers are usually larger than the small features commonly found on objects rendering them impossible to ignore.  
Fig.~\ref{compare-suction} (a) shows that a false positive when all eight vertices of the DexNet model sit on a flat surface and the spring deformations are within the threshold, but there is a groove under the suction cup gripper. Fig.~\ref{compare-suction} (b) and (e) show that there are geometries under the suction cup gripper that cause the suction cup to deform and not create proper seal.  Fig.~\ref{compare-suction} (c) shows that the DexNet model resolution is not suitable for rough surfaces. Fig.~\ref{compare-suction} (d) and (f) show that the DexNet analytical model only takes in the singulated object information. In cluttered  environments, it fails to identify neighboring objects, and the collision check performed in DexNet cannot handle these scenarios without object segmentation information.

%%%%%%%%%%%%%%%%%%%%%%%%%%%%%%%%%%%%%%%%%%%%%%%%%%%%%%%%%%%%%%%%%%%%%%%%%%%%%%%%
\section{CONCLUSIONS \& FUTURE WORK}
\vspace{-0.05in}
In this study, we have presented a novel synthetic dataset for a modular end-effector system that provides accurate grasp labels in cluttered environments with complex objects. Future work will focus on training the grasp pose detection network to obtain robust grasp pose prediction results for each individual end-effector. The end-effector with the highest prediction score, which is most likely to result in a successful grasp, will then be selected.  
Sample data from SIM-MEES can be accessed at \href{https://github.com/junchengli1/Sim-MEES.git}{https://github.com/junchengli1/Sim-MEES.git.}

%%%%%%%%%%%%%%%%%%%%%%%%%%%%%%%%%%%%%%%%%%%%%%%%%%%%%%%%%%%%%%%%%%%%%%%%%%%%%%%%
\section{ACKNOWLEDGEMENTS}
\vspace{-0.05in}
 A portion of this work was supported by a Space Technology Research Institutes grant (\# 80NSSC19K1076) from NASA’s Space Technology Research Grants Program.  
\vspace{-0.05in}

%%%%%%%%%%%%%%%%%%%%%%%%%%%%%%%%%%%%%%%%%%%%%%%%%%%%%%%%%%%%%%%%%%%%%%%%%%%%%%%%
% REFERENCES %
\bibliographystyle{IEEEtran}
\bibliography{references}

\end{document}